\documentclass[runningheads]{llncs}
\usepackage{graphicx}

\usepackage{tikz}
\usepackage{comment}
\usepackage{amsmath,amssymb} 
\usepackage{color}
\usepackage{multirow}
\usepackage{bbold}
\usepackage{bbding}

\usepackage{xcolor}

\usepackage[accsupp]{axessibility}  

\begin{document}
\pagestyle{headings}
\mainmatter
\def\ECCVSubNumber{6148}  

\title{Long-tailed Instance Segmentation using
Gumbel Optimized Loss} 

\titlerunning{Gumbel Optimised Loss}
%
\author{Konstantinos Panagiotis Alexandridis\inst{1,2}\index{Alexandridis,Konstantinos Panagiotis}\Envelope \and
Jiankang Deng\inst{3} \and
Anh Nguyen\inst{2} \and
Shan Luo\inst{1,2}
}
\authorrunning{K. P. Alexandridis et al.}
%
\institute{King's College London, London WC2R 2LS, United Kingdom \email{\{konstantinos.alexandridis,shan.luo\}@kcl.ac.uk} \and
University of Liverpool, Liverpool L69 3BX, United Kingdom \email{\{konsa15,anguyen,shan.luo\}@liverpool.ac.uk}
\and
Imperial College London, London SW7 2AZ, United Kingdom\\
\email{j.deng16@imperial.ac.uk}}

\maketitle

\begin{abstract}
Major advancements have been made in the field of object detection and segmentation recently. However, when it comes to rare categories, the state-of-the-art methods fail to detect them, resulting in a significant performance gap between rare and frequent categories. In this paper, we identify that Sigmoid or Softmax functions used in deep detectors are a major reason for low performance and are sub-optimal for long-tailed  detection and segmentation. 
To address this, we develop a Gumbel Optimized Loss ($GOL$), for long-tailed  detection and segmentation. It aligns with the Gumbel distribution of rare classes in imbalanced datasets, considering the fact that most classes in long-tailed detection have low expected probability. The proposed $GOL$ significantly outperforms the best state-of-the-art method by $1.1\%$ on $AP$, and boosts the overall segmentation by $9.0\%$ and detection by $8.0\%$, particularly improving detection of rare classes by $20.3\%$, compared to Mask-RCNN, on LVIS dataset. Code available at: \url{https://github.com/kostas1515/GOL}.
\end{abstract}

\section{Introduction}
\label{sec:intro}

There have been astonishing advancements in the fields of image classification, object detection and segmentation recently. They have been made possible by using curated and balanced datasets, e.g.,  CIFAR~\cite{krizhevsky2009learning}, ImageNet~\cite{deng2009imagenet} and COCO~\cite{lin2014microsoft} and by using deep Convolutional Neural Networks (CNNs). Despite that, all these advancements could be in vain if they are not usable in real-world applications. For example, the performance of classifiers in ImageNet is similar to humans, however, ImageNet pretrained detectors still struggle as they suffer from various sources of imbalance~\cite{oksuz2020imbalance}. Moreover, existing instance segmentation models \cite{he2017mask,cai2019cascade,chen2019hybrid} fail to generalize for long-tailed datasets and their performance significantly decreases for the rare categories \cite{gupta2019lvis}. As a result, it is difficult to exploit the advancements in image classification and transfer them to applications like object detection and segmentation due to the imbalance problem. Furthermore, there is a significant gap in performance between frequent (head) and infrequent (tail) classes in long-tailed datasets, as the state-of-the-art (SOTA) methods only detect the frequent classes \cite{wang2020devil,li2020overcoming}. All such problems may deteriorate the reliability of autonomous systems that rely on object detection and segmentation and raise concerns. 

\begin{figure}[t]
\centering
\includegraphics[scale=0.168]{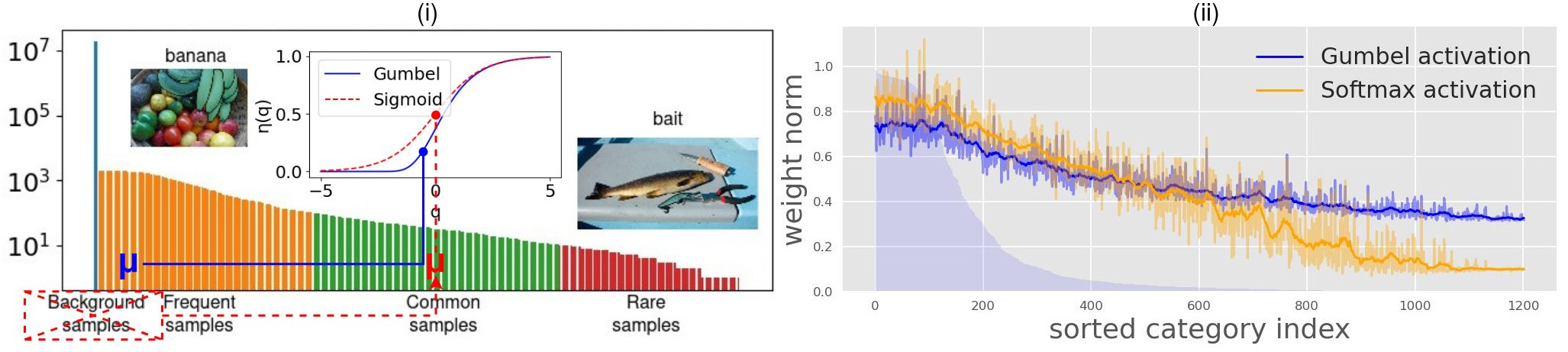}
\caption{(i) Gumbel activation function (blue) is asymmetric and it aligns better with the long-tailed instance segmentation distribution due to the extreme background and foreground class imbalance, whereas Sigmoid activation (red) is symmetric and more appropriate for balanced distributions. (ii) Gumbel activation (blue) produces more balanced weight norms in comparison to Softmax activation (orange) in the LVIS \cite{gupta2019lvis} dataset using Mask-RCNN \cite{he2017mask}.}
\label{fig:mot_wn}
\end{figure}

One possible solution for improving the long-tailed instance segmentation performance is to gather more samples for rare classes, as it is known that CNNs can achieve better results by using more data. Unfortunately, data collection will be not only costly but also intractable. The physical world contains objects that follow the Zipfian distribution \cite{liu2019large}. This means that by increasing the distinct classes of a dataset, it is unavoidable that some will be frequent while others will be rare. 

The main reason for the low performance of instance segmentation in long-tailed datasets is class imbalance. As discussed in~\cite{tan2020equalization,wang2021seesaw,mahajan2018exploring}, head classes dominate during training and they cause large discouraging gradients for tail classes. Since tail classes have fewer training samples, the amount of positive feedback is scarce and in the end, the model is trained effectively only for the head classes. It is also reflected by the norms of the classifier's weights~\cite{kang2019decoupling}: classifiers trained under the long-tailed paradigm have classification weights whose norms are larger for head classes and lower for tail classes. As larger weights produce larger probabilities, the classifiers are therefore biased towards head classes. For these reasons, many prior works focus on balancing either the weight norms or the gradients caused by head and tail classes or performing two-stage training where the model is first trained for all classes and then fine-tuned for tail categories. 

In contrast, we argue that the low performance in long-tailed instance segmentation is partially due to the use of sub-optimal activation functions in bounding box classification. Most classes in this long-tailed distribution have extremely low expected probabilities due to imbalance~\cite{oksuz2020imbalance}, making the widely used activation functions such as Sigmoid and Softmax unsuitable.
For this reason, we develop a new activation function, namely Gumbel activation and a new Gumbel loss function to model the long-tailed distribution in instance segmentation. Gumbel activation is an asymmetric function that aligns better with the long-tailed instance segmentation distribution as shown in Figure \ref{fig:mot_wn}(i). Moreover, Gumbel loss allows the gradient of positive samples to grow exponentially while suppressing the gradient of negative samples. This is especially useful for rare category learning, in which positive feedback is scarce. At the same time, it produces more balanced classification weight norms in comparison to Softmax as shown in Figure \ref{fig:mot_wn}(ii), suppressing the classification bias. Both head and tail categories can benefit from Gumbel loss, without the need of gradient re-balancing, exhausting parameter tuning, weight normalization or complex two-stage training. Furthermore, Gumbel is agnostic to frameworks, it can be used alongside with other loss functions and datasets, which makes it widely applicable.
Based on the proposed Gumbel loss, we have developed Gumbel optimized methods, that outperform the state-of-the-art instance segmentation methods on the LVIS \cite{gupta2019lvis} dataset. We list our contributions as follows:
\begin{itemize}
    \item We identify the problem of activation functions in long-tailed instance segmentation for the first time, via extensive experiments;
    \item We propose a new loss, i.e., Gumbel Optimized Loss ($GOL$), for long-tailed instance segmentation;
    \item We have validated the effectiveness of $GOL$ on real-world long-tailed instance segmentation datasets, outperforming the SOTA methods by a large margin.
\end{itemize}
\begin{figure*}[t]
    \centering
    \includegraphics[scale=0.24]{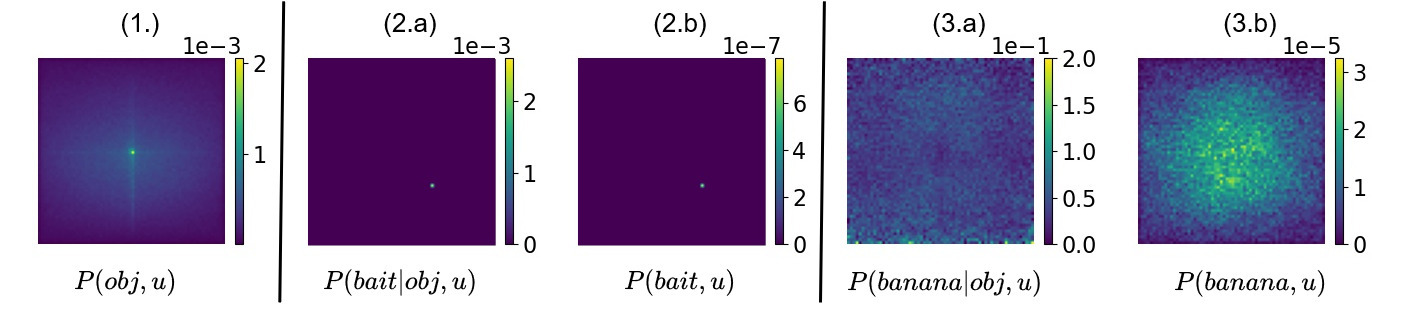}
    \caption{Object distributions in the LVIS long-tailed object detection dataset~\cite{gupta2019lvis}. (1): The distribution of objects $P(obj,u)$ in the dataset (irrespective of their class); (2.a): the class probability conditioned on object and its location $P(y|obj,u)$, and (2.b): the expected class distribution $P(y,u)$, for the tail class $bait$;
    (3.a): the class probability conditioned on object and its location $P(y|obj,u)$, and (3.b): the expected class distribution $P(y,u)$, for the head class $banana$. As shown in the figures, the distributions of objects in a long-tailed object dataset have a normal distribution as a whole and also for the head classes, whereas follows a Gumbel distribution for tail classes.}
    \label{fig:obj_det_distribution}
\end{figure*}

\section{Related works}
\label{sec:related}
\noindent \textbf{Long-tailed  object classification.} It has been a hot topic to address the imbalance problem in object classification. Long-tailed object classification datasets of CIFAR10, CIFAR100 and ImageNet have been investigated to tackle imbalanced classification using techniques such as data re-sampling~\cite{chawla2002smote,mahajan2018exploring,shen2016relay,zou2018unsupervised}, Cost Sensitive Learning (CSL)~\cite{cui2019class,khan2017cost}, margin adjustment~\cite{menon2021longtail,kim2020adjusting,cao2019learning,Ren2020balms} and two stage training~\cite{kang2019decoupling,wang2020devil}. Data
re-sampling methods re-sample the rare classes and have been most widely used and investigated~\cite{chawla2002smote,mahajan2018exploring,shen2016relay,zou2018unsupervised}. However, such methods cost more training effort and pose the risk of overfitting for rare classes, while under-sampling under-fits heads class and deteriorates the overall performance. The CSL methods construct a cost matrix so that the cost function can be more sensitive to the rare classes~\cite{cui2019class,khan2017cost}, so as to exploit the data available. But CSL methods are dependent on the dataset and require careful calibration to avoid exploding gradients caused by excessive costs. Margin adjustment techniques change the decision boundary of the classifiers by either normalizing the classifier weight norms, engineering appropriate losses or modifying the classification prediction a-posteriori~\cite{menon2021longtail,kim2020adjusting,cao2019learning,Ren2020balms}, which do not cost additional training time. Their drawback is that the margins are difficult to compute, and they are based on dataset statistics.

\noindent \textbf{Long-tailed  object detection.} Some methods addressing the long-tailed image classification could be applied in long-tailed  object detection~\cite{Ren2020balms,tan2020equalization,wang2021seesaw,kang2019decoupling}. However, many SOTA long-tailed classification methods obtain low performance for tasks that include the special \textit{background} class~\cite{mullapudi2021background}. Under this realistic scenario, the performance drop is caused by the extreme imbalance between the dominant \textit{background} class and other foreground classes. 
The same applies in long-tailed  object detection where the imbalance factor is $\sim 1000$ larger than the imbalance factor in image classification. For this reason, not all long-tailed  classification methods are transferable to long-tailed  instance segmentation. Instead, many methods are developed to tackle long-tailed  object detection, directly. Some of them include the creation of specialized loss functions that balance the gradient contribution of positive and negative samples ~\cite{tan2020equalization,tan2021equalization,wang2021adaptive,wang2021seesaw,peng2020large}. Others construct hierarchical groups \cite{li2020overcoming,wu2020forest}, enforce margins in the classifier's prediction \cite{feng2021exploring,Ren2020balms,tang2020long} or use two-stage strategies \cite{wang2020devil,kang2019decoupling,zhang2021distribution}. These methods have produced promising results but they suffer from limitations: Two-stage methods are complex and laborious; hierarchical methods require pre-processing and careful grouping; loss engineering methods have many hyper-parameters that need tuning. All these methods use Sigmoid or Softmax as their activation function, which is not close to the target distribution and not a good choice as we discuss in Section~\ref{sec:problem}. 

To the best of our knowledge, we are the first to tackle long-tailed segmentation by using Gumbel loss function.
The most related work is \cite{bridge2020introducing}, where they used the general extreme value distribution to classify Covid-19 cases. In contrast, we develop Gumbel for long-tailed instance segmentation. 

\section{Problem Formulation}
\label{sec:problem}

Assume a dataset $X = \{x_i,y_i\},i\in\{1,...,N\}$, where $x_i,y_i$ are images and annotations respectively and $N$ is the total number of training images. We can train a convolutional neural network $f(X,\theta)=z$, where $z$ is the latent representation and $\theta$ is the network's weight parameters. To calculate the prediction $\bar{y}$, one first can use a fully connected layer $q(z)=W^Tz +b$, where $W$ is the classification weights and $b$ is the bias term, to calculate the score $q_i$. Then $\bar{p_i}=\eta(q_i)$ is used to transform the score $q_i$ into probability $\bar{p_i}$, using the activation function $\eta(\cdot)$ and finally the prediction $\bar{y}$ is calculated using $\bar{y}=\arg \max_i(\bar{p_i})$.

In image classification and instance segmentation, the Sigmoid activation, i.e.,  $\eta_{\text{sigmoid}}(q_i)=\frac{1}{1+e^{-q_i}}$, or the Softmax activation, i.e.,  $\eta_{\text{softmax}}(q_i)=\frac{e^{q_i}}{\sum{e^{q_j}}}$, has been commonly used. For the binary case, it assumes that $\bar{p_i}$ follows a Bernoulli distribution as the score  $q_i=\text{log}\frac{\bar{p_i}}{1-\bar{p_i}}$ and it can be interpreted as the odds-ratio of the event $\bar{p_i}$, i.e., how many times an event happens $\bar{p_i}$ divided by how many times it does not happen $1-\bar{p_i}$ in a log scale.
 
 
It would be reasonable to use the Sigmoid or Softmax activation function for image classification, where the expected probability distribution $\mathbb{P}$ is a Bernoulli distribution and all classes are mutually exclusive, thus one can use Sigmoid or Softmax to effectively model the data.
However, we argue that it would not be well suitable to use these activation functions for long-tailed instance segmentation as the expected distribution of objects is not the same as the expected image distribution in classification. Object distribution is more complex as it is affected by class imbalance and location imbalance.  

\subsection{Object Distribution}
\label{obj_distr_calculation}
To make this clear, we first calculate the ground truth object distribution. To this end, we calculate the expected number of objects $P(obj,u)$ whose centers fall inside the cell $u=[i,j]$ of the normalized grid as follows:
\begin{equation}
    P(obj,u)=\frac{\sum_{x=1}^{x=N}\mathbb{1}(obj,x) \mathbb{1}(obj,u)}{M}
    \label{pobj}
\end{equation}
where $obj$ is the object occurrence, $\mathbb{1}$ is the indicator function and $M$ is the total number of objects in the dataset.
Next, we calculate the class membership $P(y|obj,u)$ for each location $u$, which summarizes the uncertainty of an object belonging to each class $y$ in the dataset for the specific location $u$ (i.e., it holds that $ \sum_{y=1} ^{y=C}{P(y|obj,u)=1}$). Finally, we calculate $P(y,u)$\footnote{Here we omit $obj$ for simplicity since $P(y,obj,u) = P(y,u)$ due to that $y$ shows there is object occurrence $obj$.} as:
\begin{equation}
    P(y,u) = P(y|obj,u)P(obj,u)
    \label{eq:obj_distr}
\end{equation}

The final target distribution is a distribution that we aim to estimate by minimizing the cross entropy between the target and the data distribution.

In Figure \ref{fig:obj_det_distribution}, we visualize $P(obj,u)$, $P(y|obj,u)$ and $P(y,u)$ for one head class and one tail class of LVIS dataset. 
The probability of detection for a head class i.e.,  \textit{banana} in the example, is low in each location and varies in different locations of the image. For a tail class, i.e.,  \textit{bait} in the example, is even lower and zero for most locations. This is different from long-tailed  image classification, in which the expected class probabilities are not affected by location imbalance, only by class imbalance. On the other hand, in long-tailed  instance segmentation, classes have even lower expected probabilities as they are affected by both location imbalance and class imbalance. For example, even head classes like \textit{banana} have even lower expectation for locations far away from the center of the image, making the long-tailed  segmentation more challenging for both head and tail classes. This highlights the magnitude of imbalance in long-tailed  segmentation and motivates us to develop Gumbel activation.

Using Gumbel, we assume that the target distribution follows Gumbel distribution and this is a better choice than using Sigmoid or Softmax because the expected classification probabilities are minuscule. In fact, by using any activation function, one assumes how the ground truth is distributed. It is a common practice to use Sigmoid or Softmax and this assumes that the target distribution is Bernoulli. While this is a rational choice for image classification, it is unrealistic for long-tailed instance segmentation as the expected classification probabilities are infinitesimal. For this reason, we assume that the target distribution is Gumbel and we use Gumbel activation. We further explain why choosing an activation, that implicitly assumes the target distribution, below.

\subsection{Activations as Priors} 
\label{subsec:activ_as_prior}
To understand why choosing an activation implicitly assumes the target distribution, we consider an example of binary classification, however, it can be extended to multi-class classification easily. In a binary classification problem, the true variable $y$ relates with the representations $z$ as follows:
\begin{equation}
y=
    \begin{cases}
        1, \;\; \text{if} \;\;  W^Tz +b + \epsilon>0\\
        0, \;\; \text{otherwise}
    \end{cases}
\end{equation}
where $\epsilon$ is the error that is a random variable. The classification boundary is set to 0, but it could be any other value as it is adjusted by the bias term $b$ in optimization.
We are interested in the probability $P(y=1)$ and this is calculated as:
\begin{equation}
\begin{aligned}
P(y=1)=P(W^Tz+b + \epsilon>0) \\
P(\epsilon>-W^Tz-b)=1-F(-q)
\label{eq:cumulative}
\end{aligned}
\end{equation}
where $F$ is the cumulative distribution function.
Many practitioners use Sigmoid to activate $q$ and estimate $P(y=1)$ and this means that:

\begin{equation}
\begin{aligned}
P(y=1)=\eta_{\text{sigmoid}}(q)= \frac{1}{1+e^{-q}} \\
\eta_{\text{sigmoid}}(q) =F_{\text{logistic}}(q;0,1) \\ P(y=1)=1-F_{\text{logistic}}(-q;0,1)
\label{eq:log_prior}
\end{aligned}
\end{equation}

By comparing Eq. \ref{eq:cumulative} and the last expression of Eq. \ref{eq:log_prior}, it is understood that by using Sigmoid activation, one assumes that the error term $\epsilon$ follows the standard logistic distribution (with $\mu=0$ and $\sigma=1$), as one chooses $F$ to be logistic.
In practice, when any activation function $\eta(q)$ is applied, it is implicitly assumed how $\epsilon$ is distributed and as a result the target distribution $y\sim\mathbb{P}$ is assumed. For example, if a Sigmoid function is used, the variable $q$ is transformed into a binomial probability distribution, which implies that it has a success rate $\bar{p}$ and a failure rate $1-\bar{p}$, like a coin toss. In this case, it is assumed that $y$ follows Bernoulli Distribution and the error $\epsilon$ follows Logistic Distribution. Finally, the training of the model is to minimize the discrepancy between the target distribution $\mathbb{P}$ and the predicted distribution $\mathbb{Q}$ using Cross Entropy:
\begin{equation}
    H(\mathbb{P}(x),\mathbb{Q}(x))= -\sum_{x \in X} \mathbb{P}(x) \text{log}(\mathbb{Q}(x))
    \label{cross_entropy}
\end{equation}
In training, as Stochastic Gradient Descent is an iterative algorithm, the starting conditions play a significant role. This suggests that it is preferable to have a good starting prior so that the initial estimation of $\mathbb{P}$ is reasonable and the choice of the activation function will facilitate the initial estimation. This has a similar concept to the prior distribution in Bayesian Inference, where it is important to choose a suitable prior for optimal results. 


\noindent\textbf{Hypothesis.} In long-tailed instance segmentation, the expected classification probabilities are significantly low due to imbalance problems as mentioned by \cite{oksuz2020imbalance} and explained in Eq.~\ref{eq:obj_distr}. For this reason, we hypothesize that Gumbel activation will produce superior results as it models the data using Gumbel Distribution which is closer to the real object distribution.

In conclusion, long-tailed instance segmentation is far more challenging than image classification and naive usage of Sigmoid or Softmax activation infringes on the underlying assumptions. 
Nevertheless, one can use Cross Entropy  to minimize the discrepancy between the target distribution $\mathbb{P}$ and the predicted distribution $\mathbb{Q}$
but if the prior guess is significantly different than the actual distribution $\mathbb{P}$ then the results might be sub-optimal. By choosing the activation function, one guesses how the target $\mathbb{P}$ is distributed. For Sigmoid or Softmax, one believes that $\mathbb{P}$ is Bernoulli distribution and while this is reasonable for image classification, we argue that in long-tailed instance segmentation it is not optimal and we empirically show that Gumbel can produce superior results.
\section{Gumbel Activation for Long-tailed Detection}
\label{sec:methodology}

\subsection{Sigmoid Activation for Object Classification}
It is useful to remind the readers about Sigmoid activation as we can make clear distinctions between this and our suggested activation. The formula is  $\eta_{\sigma}(q_i)=\frac{1}{1+\exp(-q_i)}$.
If one encodes the ground truth $y$ as a one-hot vector then the gradient using Eq.~\ref{cross_entropy} is $\frac{dL(\eta_{\sigma}(q_i),y_i)}{dq_i}=y_i(\eta_{\sigma}(q_i)-1)+(1-y_i)\eta_{\sigma}(q_i)$.

Sigmoid is a symmetric activation function: the positive gradient (i.e.,  when $y=1$) takes values from $(-1,0)$, while the negative gradient (i.e.,  when $y=0$) takes values from $(0,1)$ and the response value grows with the same rate for both positive and negative input, as shown in Figure \ref{fig:activation_summary_figures}(i).

\subsection{Gumbel Activation for Rare-class Segmentation}
\label{gumbel_methodology}
We notice that $P(y,u)$ has infinitesimal probabilities. For tail classes it has a maximum value in the scale of \textit{1e-7}, for head classes it has a maximum \textit{1e-5} and for both cases, it has even lower probabilities for edge locations in the image. This motivates us to model $P(y,u)$ using the extreme value distribution Gumbel. Gumbel is useful for modelling extreme events, i.e., those with very low probabilities. For example, Gumbel can be used to predict the next great earthquake because this has a much lower probability than regular earthquakes. Other applications of Gumbel can be found in finance and biology and the readers are referred to this work \cite{kotz2000extreme} for more information on extreme value distribution. In long-tailed  instance segmentation, one can use the cumulative density function of the standard Gumbel distribution as the activation function (we further study the choice of non-standard Gumbel activation in supplementary material):
\begin{equation}
    \eta_{\gamma}(q_i)=F_{\text{Gumbel}}(q;0,1)=\exp(-\exp(-q_i))
    \label{gumbel_activation}
\end{equation}
Combining Eq.~\ref{cross_entropy} and Gumbel activation we derive Gumbel Loss (GL) as:
\begin{equation}
    GL(\eta_{\gamma}(q_i),y_i)= 
    \begin{cases}
        -\log(\eta_{\gamma}(q_i)),\;\; if \;\;  y_i=1\\
        -\log(1-\eta_{\gamma}(q_i)),\;\;  if\;\;  y_i=0
    \end{cases}
    \label{gumbel_loss}
\end{equation}

\begin{figure}
    \includegraphics[scale=0.13]{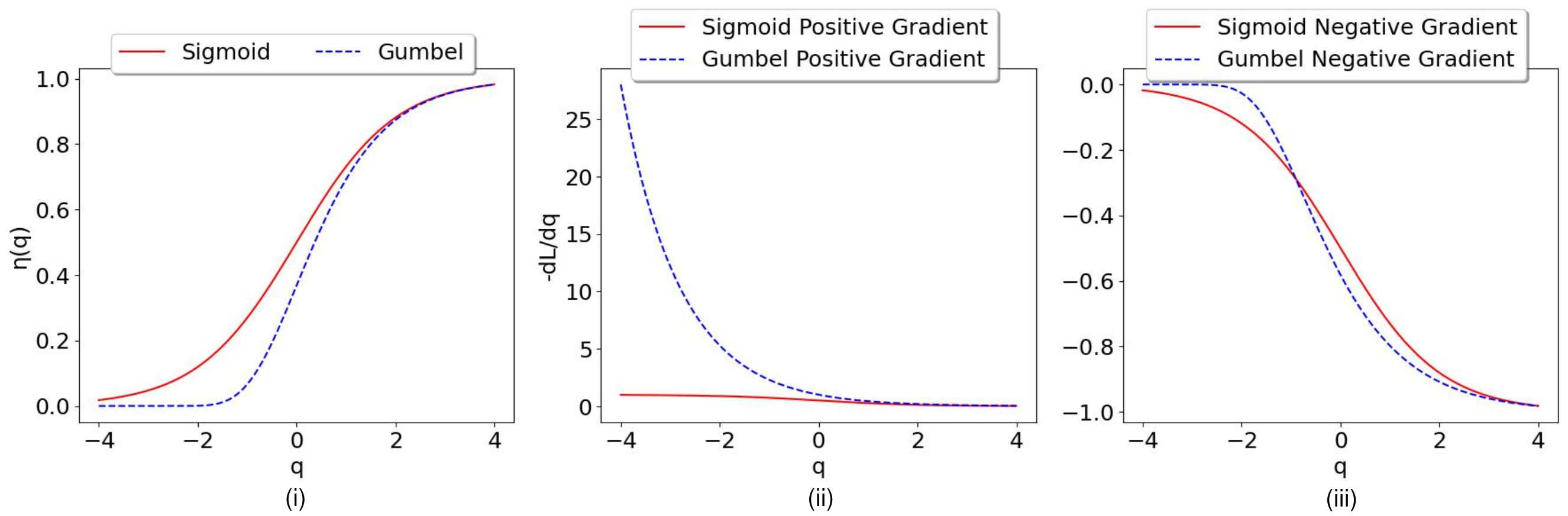}
    \caption{Activation properties for Gumbel (blue dashed lines), Sigmoid (red solid lines) activation functions. Their activation behaviours are illustrated in (i). The Gumbel activation is asymmetric, whereas Sigmoid is symmetric. Their positive gradients and negative gradients are illustrated in (ii) and (iii) respectively, using inverted y-axis $\frac{-dL}{dq}$. The positive gradients of the Gumbel activation (i.e.,  when $y=1$) ranges in $(-\infty,0)$, while the negative gradients (i.e.,  when $y=0$) ranges in $(0,1)$.}
    \label{fig:activation_summary_figures}
\end{figure}

\noindent The gradient of Eq.~\ref{gumbel_loss} is:
\begin{equation}
    \frac{dL(\eta_{\gamma}(q_i),y_i)}{dq_i}= 
    \begin{cases}
        -\exp(-q_i),\;\; if \;\;  y_i=1\\
        \frac{\exp(-q_i)}{\exp(\exp(-q_i))-1},\;\;  if\;\;  y_i=0
    \end{cases}
    \label{gumbel_gradient}
\end{equation}
The Gumbel activation is asymmetric as illustrated in Figure \ref{fig:activation_summary_figures}(i). This means that the positive gradients (i.e.,  when $y=1$) will take values from $(-\infty,0)$ while the negative gradients (i.e.,  when $y=0$) will take values from $(0,1)$. This is a beneficial property that allows the positive feedback to grow exponentially while it suppresses the negative feedback. It is especially useful in long-tailed segmentation as the positive feedback for rare categories is scarce. With Gumbel activation, positive gradients can grow faster than by using Sigmoid activation as shown in Figure \ref{fig:activation_summary_figures}(ii) and this can boost the performance of rare classes.

\subsection{Gumbel Optimised Loss}
Gumbel activation can be used not only with Cross Entropy loss but with other state-of-the-art loss functions as well. We select the recently proposed Droploss \cite{hsieh2021droploss} as the loss function of our enhanced Gumbel-based model, other loss functions are shown in Fig. \ref{fig:backbones_losses}(ii). Using Gumbel activation and Droploss we propose Gumbel Optimised Loss $GOL$ as follows:
\begin{equation}
    \mathcal{L}_{GOL} = -\sum_{j=1}^{C}w_j^{Drop}\log(\bar{p_j}),\;\; \bar{p_j}=\begin{cases}
         \eta_{\gamma}(q_i),\;\; if \;\;  y_j=1\\
         1-\eta_{\gamma}(q_i),\;\;  if\;\;  y_j=0
        \end{cases}
\end{equation}
where $w_j^{Drop}$ are class specific weights proposed by DropLoss \cite{hsieh2021droploss}. We show the full equation in supplementary material due to space limitations.

\section{Experimental Setup}
\label{sec:experimental_setup}

\noindent\textbf{Dataset and evaluation metrics.}
For our experiments on long-tailed instance segmentation, we use LVIS (Large Vocabulary Instance Segmentation) dataset \cite{gupta2019lvis}. We mainly use version 1 which contains 100k images for training and 19.8k images for validation. LVISv1 dataset contains $1,203$ categories that are grouped according to their image frequency: \textit{rare} categories (those with 1-10 images in the dataset), \textit{common} categories (11 to 100 images in the dataset) and \textit{frequent} categories (those with $>100$ images in the dataset.) Some previous methods use the LVISv0.5 dataset, which has 1230 classes instead of 1203.  For fairness, we also show results in this dataset. 
We report our results using average segmentation performance $AP$, average box performance $AP^b$ and average segmentation performance for rare $AP^r$, common $AP^c$ and frequent categories $AP^f$. 


\noindent\textbf{Implementation details.}
For our experiments, we use 4 V100 GPUs, a total batch size of 16 images, a learning rate of 0.02, weight decay of 0.0001, Stochastic Gradient Descent and momentum of 0.9. We use random horizontal flipping and multi-scaling as data augmentations for training, following the conventions of the community. We train our models using the \textit{mmdetection} framework \cite{mmdetection} and during inference, we use score threshold of 0.0001 as in \cite{gupta2019lvis}.

For our intermediate experiments, we use Gumbel activation and a plethora of architectures, backbones, loss functions and sampling strategies using the 1x schedule. For our enhanced $GOL$ model, we use Mask-RCNN \cite{he2017mask}, the 2x schedule, Normalised Mask \cite{wang2021seesaw}, RFS sampler \cite{gupta2019lvis} and a stricter Non-Maximum Suppression threshold that is 0.3. When using Gumbel activation, we initialize the weights of the classifier to 0.001 and the bias terms to -2.0, to enable stable training. More implementation details, design choices and results are discussed in our supplementary material.


\section{Results}
\label{sec:experiments}

\begin{table*}[t]
\centering
\caption{Comparative results for LVISv1 using schedule 1x, random sampler (left) and RFS \cite{gupta2019lvis} sampler (right). Gumbel activation is superior than Softmax, especially for the case of random sampling when the distribution is unaltered.}
\label{my-label}
\begin{tabular}{l|l|l|l|l|l||l|l|l|l|l|l}

Sampler &
\multicolumn{5}{c||}{Random}&\multicolumn{6}{c}{RFS \cite{gupta2019lvis}}\\
\hline
Method &$AP$&$AP^r$&$AP^c$&$AP^f$&$AP^b$&Method &$AP$&$AP^r$&$AP^c$&$AP^f$&$AP^b$ \\
\hline
Sigmoid&16.4 &0.8 &12.7 &27.3 &17.2 & Sigmoid&22.0 &11.4 &20.9 &27.9 &\textbf{23.0}\\
Softmax &15.2 & 0.0& 10.6 &26.9&16.1& Softmax &21.5 & 9.7& 20.7 &27.6&22.4\\
Gumbel (ours) & \textbf{19.0}&\textbf{4.9}&\textbf{16.8}&\textbf{27.6}&\textbf{19.1}&Gumbel (ours) & \textbf{22.7}&\textbf{12.2}&\textbf{21.2}&\textbf{28.0}&22.9\\

\end{tabular}
\label{tab:samplers}
\end{table*}



\subsection{Results with Different Sampling Strategies}
We compare activations when using a random sampler and 1x schedule. Under this configuration, the target distribution is unaltered and the probability of sampling is equal to the dataset's class probability.
As Table \ref{tab:samplers} suggests, the best activation function using a random sampler is Gumbel which largely outperforms Sigmoid and Softmax. It increases overall $AP$ by 2.6\%, $AP^r$ and $AP^c$ by 4.1\% and $AP^b$  by 1.9\% compared to Sigmoid and $AP$ by 3.8\%, $AP^r$ by 4.9\%, $AP^c$ by 6.2 \% and $AP^b$ by 3.0\% compared to Softmax. Noticeably, the gap in mask and box performance is smaller with Gumbel activation at 19.0\% and 19.1\% respectively, while other activations have larger gaps between box and segmentation performance. This suggests that Gumbel is more suitable for long-tailed segmentation than other activation functions.

We apply the state-of-the-art RFS \cite{gupta2019lvis} method, which is an oversampling method and report the results in Table \ref{tab:samplers}. With RFS, images containing rare categories are up-sampled, thus the original distribution is distorted. Under this scenario, Gumbel activation does not boost the performance as much as before, as the object distribution becomes more balanced with oversampling. Nevertheless, Gumbel improves overall performance by a respective margin of 0.7\% in overall segmentation and by 0.8\% $AP^r$ which is attributed to the fact that firstly, there are still classes in the dataset that suffer from imbalance after using RFS, thus Gumbel can model them better than Sigmoid or Softmax; secondly, Gumbel activation has a lower gap in bounding box and segmentation performance than Sigmoid or Softmax, which results in higher segmentation performance (i.e., 0.7\% increase) given similar box performance.
\begin{figure}[t]
    \centering
    \includegraphics[scale=0.19]{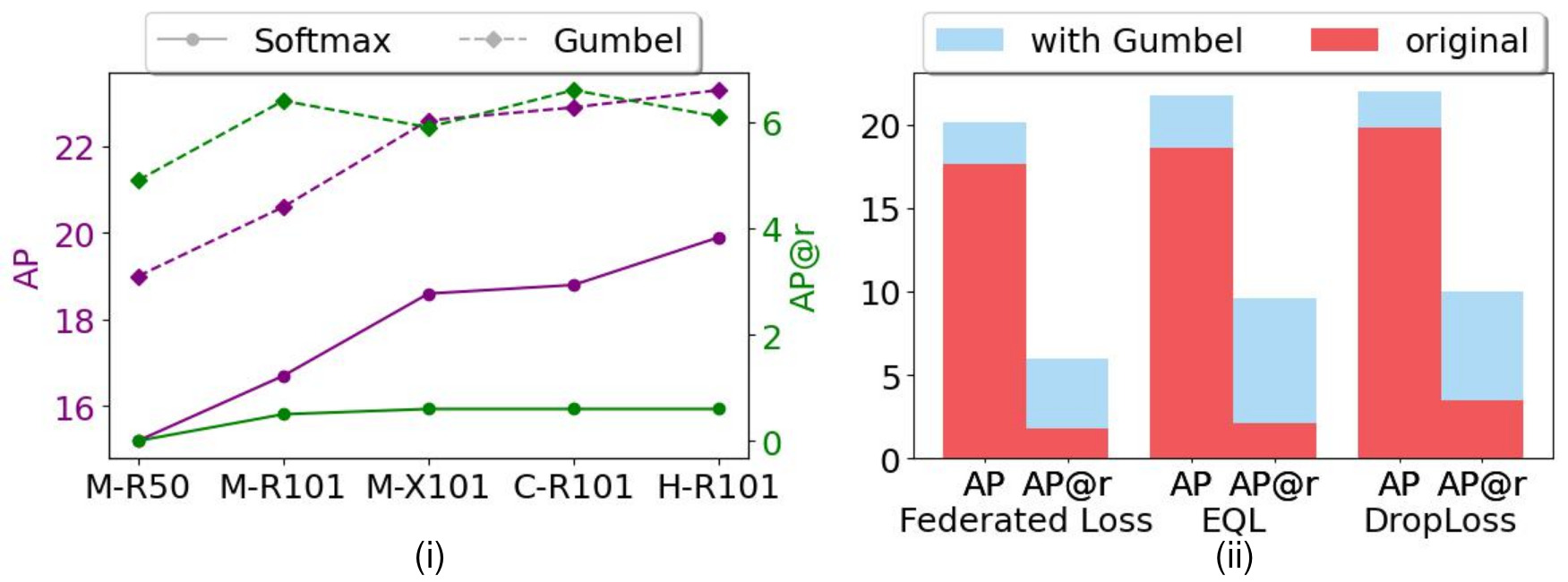}
    \caption{(i): Comparison of Softmax against Gumbel activation using common instance segmentation frameworks. (ii): Comparison of Sigmoid-based SOTA losses against the Gumbel-based alternatives. All models use 1x schedule and random sampling.}
    \label{fig:backbones_losses}
\end{figure}

\subsection{Integrating Gumbel Activation}
We conduct experiments using training schedule 1x with larger backbones, i.e., Resnet101 \cite{he2016deep} and ResNeXt101 \cite{xie2017aggregated} and architectures i.e., Cascade Mask RCNN \cite{cai2019cascade} and Hybrid Task Cascade \cite{chen2019hybrid} to determine if the proposed activation generalizes for deeper models. As shown in Figure \ref{fig:backbones_losses} (i), Gumbel activation is a better choice than the Softmax activation function as it achieves better overall AP and AP in rare categories. 

Next, we examine the behavior of the Gumbel activation function using SOTA loss functions Equalization loss (EQL) \cite{tan2020equalization}, DropLoss \cite{hsieh2021droploss} and Federated Loss \cite{zhou2021probablistic}. For EQL and DropLoss, we use the hyperparameter $\lambda=0.0011$ which is more appropriate for LVISv1 as described in \cite{tan2021equalization} and we change only the activation function from Sigmoid to Gumbel. For Federated Loss, we use the same hyper-parameters as described in \cite{zhou2021probablistic} changing only the activation from Sigmoid to Gumbel. As Figure  \ref{fig:backbones_losses} (ii) indicates, Gumbel significantly boosts the performance of all models in both overall AP and rare category AP and this highlights its applicability and efficacy. 
We show more detailed results in our supplementary material.



\subsection{GOL Components}
We conduct an ablation study using a 2x-schedule and we report the most significant findings, a more detailed ablation study is provided in our supplementary material.
We use the standard Mask-RCNN, EQL \cite{tan2020equalization} and RFS \cite{gupta2019lvis} as the basis and examine the behavior of Gumbel.
As shown in Table \ref{tab:ablation}, Gumbel can significantly boost the performance of this pipeline by $0.8\%$. To further boost the performance, we use a stricter Non-Maximum Suppression threshold that is $0.3$ and Mask normalization, we denote these enhancements in the Table as (Enh). Next, we adopt DropLoss which is a recent improvement in the loss function of EQL, proposed by \cite{hsieh2021droploss}.
Finally, the best performance is achieved using RFS, Gumbel, Enh and DropLoss, we codename this as pipeline $GOL$. 

\begin{table}
    \centering
    \caption{Ablation study, using Mask-RCNN, Resnet50 and training schedule 2x.}
    \begin{tabular}{ccccc|c|c|c|c|c}

         RFS&EQL&Gumbel&Enh&DropLoss&$AP$ &$AP^r$  &$AP^c$  &$AP^f$  &$AP^b$  \\
         \hline
         &&&&&18.7&1.1&16.2&29.2&19.5\\
         \checkmark&&&&&23.7&13.3&23.0&29.0&24.7\\
         \checkmark&\checkmark& &&&25.3&17.4&24.9&29.2&26.0\\
         \checkmark&\checkmark&\checkmark&&&26.1&18.4&25.9&29.8&26.8\\
         \checkmark&\checkmark&\checkmark&\checkmark&&26.9&18.1&26.5&\textbf{31.3}&26.8\\
         \checkmark&&\checkmark&\checkmark&\checkmark&\textbf{27.7}&\textbf{21.4}&\textbf{27.7}&30.4&\textbf{27.5}\\
    \end{tabular}
    \label{tab:ablation}
\end{table}

\noindent\textbf{Total performance.} In the end, our $GOL$ method significantly improves the vanilla Mask-RCNN $AP$ by $9.0\%$, and it largely improves $AP^r$ by $20.3\%$, $AP^c$ by $11.5\%$, $AP^f$ by $1.2\%$ and $AP^b$ by $8.0\%$.

\begin{figure*}[t]
    \centering
    \includegraphics[scale=0.4]{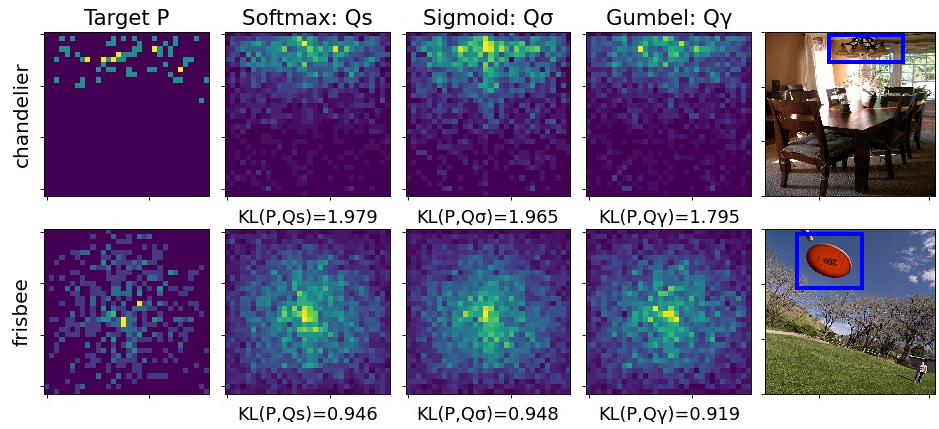}
    \caption{Comparison of two object distributions in LVIS validation set, using Softmax (second column), Sigmoid (third column) and Gumbel (fourth column). Gumbel predicts distributions that have smaller KL divergence than Sigmoid or Softmax.}
    \label{fig:distance}
\end{figure*}

\noindent\textbf{Comparison with other methods.}
As shown in Table \ref{tab:sota}, $GOL$ significantly surpasses the state-of-the-art in LVIS Dataset using the standard Mask-RCNN benchmark. In detail, in LVISv0.5, $GOL$ achieves 29.5\% $AP$, surpassing RFS \cite{gupta2019lvis} by 4.1\% and the best state-of-art LOCE \cite{feng2021exploring} by 1.1\% in $AP$. Moreover, our method achieves the best performance at rare and common categories and it consistently surpasses many other recent works.
In LVISv1, $GOL$ achieves 27.7\% overall $AP$, surpassing RFS\cite{gupta2019lvis} by 4.0\%, LOCE by 1.1\%, Seesaw Loss \cite{wang2021seesaw} by 1.3\% and EQLv2\cite{tan2021equalization} by 2.2\% using ResNet50 backbone. It also achieves the best $AP$, which is at least 0.9\% higher than other methods, using the larger ResNet101 backbone. Finally, it consistently outperforms all other methods for rare and common categories in both Resnet50 and Resnet101 backbones. 

\begin{table*}
    \centering
        \caption{Comparison against SOTA on the LVIS dataset.}
        \begin{tabular}{c|c|c|ccccc}
        Method&Dataset&Framework &$AP$&$AP^r$&$AP^c$&$AP^f$&$AP^b$ \\
         \hline
        RFS \cite{gupta2019lvis}&\multirow{ 9}{*}{LVIS v0.5}&\multirow{ 9}{*}{Mask-RCNN R50-FPN}&25.4&16.3&25.7&28.7&25.4 \\
        DropLoss\cite{hsieh2021droploss}&&&26.4&17.3&28.7&27.2&25.8 \\
        BAGS \cite{li2020overcoming}& & &26.2&18.0&26.9&28.7&25.8 \\
        BALMS\cite{Ren2020balms}& & &27.0&19.6&28.9&27.5&27.6\\
        Forest-RCNN\cite{wu2020forest}& & &25.6&18.3&26.4&27.6&25.9 \\
        EQLv2\cite{tan2021equalization}& & &27.1&18.6&27.6&29.9&27.0\\
        LOCE\cite{feng2021exploring}& & &28.4 &22.0&29.0&30.2&\textbf{28.2}\\
        DisAlign \cite{zhang2021distribution}& & &27.9 &16.2&29.3&\textbf{30.8}&27.6\\
        $GOL$ (ours) & & &\textbf{29.5} &\textbf{22.5}&\textbf{31.3}&30.1&\textbf{28.2}\\
        \hline
        RFS\cite{gupta2019lvis}&\multirow{6}{*}{LVIS v1.0}&\multirow{6}{*}{Mask-RCNN R50-FPN}&23.7&13.3&23.0&29.0&24.7\\
         EQLv2\cite{tan2021equalization}&&&25.5&17.7&24.3&30.2&26.1\\
          LOCE\cite{feng2021exploring}& & &26.6 &18.5 &26.2&\textbf{30.7}&27.4\\
         NorCal with RFS \cite{pan2021model}&&&25.2&19.3&24.2&29.0&26.1\\
         Seesaw\cite{wang2021seesaw}&&&26.4 &19.5&26.1&29.7&\textbf{27.6}\\
         $GOL$ (ours)&& &\textbf{27.7}&\textbf{21.4}&\textbf{27.7}&30.4&27.5\\
         \hline
         RFS \cite{gupta2019lvis}&\multirow{6}{*}{LVIS v1.0}&\multirow{6}{*}{Mask-RCNN R101-FPN}&25.7&17.5&24.6&30.6&27.0\\
         EQLv2\cite{tan2021equalization}&&&27.2&20.6&25.9&31.4&27.9\\

         LOCE\cite{feng2021exploring}&&&28.0 &19.5 &27.8&\textbf{32.0}&29.0\\
         NorCal with RFS \cite{pan2021model}&&&27.3&20.8&26.5&31.0&28.1\\
         Seesaw\cite{wang2021seesaw}&&&28.1&20.0&28.0&31.8&28.9\\
         $GOL$(ours)&&&\textbf{29.0}&\textbf{22.8}&\textbf{29.0}&31.7&\textbf{29.2}\\
    \end{tabular}
    \label{tab:sota}
\end{table*}

\subsection{Model Analysis}
We analyze the behavior of Mask-RCNN using Gumbel Loss. In detail, we visualize Mask-RCNN predicted object distributions in the validation set, for two random classes \textit{chandelier} and \textit{frisbee}. We compare, the predicted distributions $\mathbb{Q}$ of Softmax, Sigmoid and Gumbel against the ground truth $\mathbb{P}$ using the Kullback Leibler divergence.

As Figure \ref{fig:distance} suggests, Gumbel produces object distributions that are closer to the target distribution, as they have smaller Kullback–Leibler (KL) divergence.

Moreover, as \cite{kang2019decoupling} has suggested, there is a positive correlation between the weight norms of the classifier and the image frequency of categories, which results in classification bias. In our case, we visualize the weight norms of the Mask-RCNN classifier trained with Softmax (baseline model), and Gumbel respectively. As Figure \ref{fig:mot_wn} (ii) suggests, the weight norm distribution of the Mask-RCNN classifier trained with Gumbel is more uniform than the distribution of vanilla Mask-RCNN. This suggests that the classifier norms are more balanced when using Gumbel loss which validates its efficacy.

\subsection{Long-tailed Image Classification}
We further test Gumbel activation in long-tailed image classification benchmarks. For all classification experiments, we use random sampling and decoupled strategy. In decoupled strategy, the model is first trained with Softmax activation and then only the classifier is re-trained with Gumbel activation.

\noindent\textbf{CIFAR100-LT} \cite{cao2019learning}. We train a ResNet32 for 240 epochs using Auto-Augment \cite{cubuk2019autoaugment}, SGD, weight decay 0.0002, batch size 64 and learning rate 0.1 that decays at epoch 200 and 220 by 0.01. In the second stage, we retrain the classifier using a learning rate of 1e-4 for 15 epochs.

\noindent\textbf{ImageNet-LT, Places-LT} \cite{liu2019large}. For ImageNet-LT, we train ResNet50 with and without Squeeze and Excite \cite{hu2018squeeze} modules for 200 epochs using Auto-Augment. For Places-LT, we finetune an ImageNet pretrained ResNet152 for 30 epochs. For both datasets, we use SGD, weight decay 0.0005, batch size 256, and learning rate 0.2 with cosine scheduler. In the second stage, we retrain the classifier using a learning rate of 1e-5 for 10 epochs. 

\begin{table}
    \centering
    \caption{Top-1 accuracy on long-tailed classification datasets.}
    \begin{tabular}{c|ccc|cc|c}
         Dataset&\multicolumn{3}{c|}{CIFAR100-LT}&\multicolumn{2}{c|}{ImageNet-LT}&Places-LT\\
         \hline
         Imbalance factor&50&100&200&\multicolumn{2}{c|}{256}&996\\
         Model&\multicolumn{3}{c|}{ResNet-32}&ResNet-50&SE-ResNet-50&ResNet-152\\
         
         \hline
         Softmax &46.2&42.4&38.3&45.2&45.9&28.7\\
         Gumbel &\textbf{49.0}&\textbf{45.5}&\textbf{41.5}&\textbf{48.2}&\textbf{48.5}&\textbf{30.0}\\
    \end{tabular}
    \label{tab:classif_res}
    \vspace{-4mm}
\end{table}

As Table \ref{tab:classif_res} indicates, Gumbel activation can boost the classification performance of all models in all datasets consistently.

\section{Conclusion and Discussions}
\label{conclusion}
 We hypothesize that real-world long-tailed detection and segmentation data follows a distribution that is closer to Gumbel distribution and not Bernoulli. For this reason, we propose to use Gumbel activation instead of Sigmoid or Softmax. We validate the superiority of Gumbel against Sigmoid and Softmax under different sampling strategies, deeper models and loss functions and we develop the $GOL$ method based on Gumbel activation that significantly outperforms the state-of-the-art. Our extensive experiments validate that Gumbel is a superior activation function that can be used as a component with both off-the-shelf methods and state-of-the-art models to further increase their performance.
 
 We have also tested Gumbel activation in long-tailed classification benchmarks and saw consistent improvements when Gumbel is used as a decoupled method. Finally, Gumbel could also be used for dense object detection and we have seen a $0.4\%$ increase in AP when using RetinaNet on COCO and 1x schedule.
 Currently, Gumbel cannot be used with Softmax-based loss functions and it does not take full advantage of oversampling methods. In the future, we will develop a custom loss function and sampling mechanism tailored to Gumbel activation. 
 
\hfill \break
\noindent\textbf{Acknowledgments}
This work was supported by the Engineering and Physical Sciences Research Council (EPSRC) Centre for Doctoral Training in Distributed Algorithms [EP/S023445/1]; EPSRC ViTac project (EP/T033517/1); King's College London NMESFS PhD Studentship; the University of Liverpool and Vision4ce. It also made use of the facilities of the N8 Centre of Excellence in Computationally Intensive Research provided and funded by the N8 research partnership and EPSRC  [EP/T022167/1]. 
%
%

\clearpage
\appendix

\section{Gumbel activation}
\label{gumbel_details}
\subsection{Weights and biases initialization}
Gumbel activation has exponential positive gradients, making it difficult to initialize due to arithmetic errors caused by the gradient overflow. For this reason, one should initialize the bias and weight terms of the classification layer with values that will produce small initial gradient. First, all weight terms $W^T$ are initialized to a small value of $0.001$, which will result in that all $q_i=W^Tz+b \approx b$, then the total gradient will be:
\begin{equation}
    \nabla H(\eta_{\gamma}(q),y)\approx -\exp(-b)+(C-1)\frac{\exp(-b)}{\exp(\exp(-b))-1}
    \label{eq:weight_int_1}
\end{equation}
where $C$ is the total number of classes in the dataset. As the total gradient should be zero initially, we have:
\begin{equation}
    \begin{aligned}
     \nabla H(\eta_\gamma(q),y)=0 \\
  (C-1)\frac{\exp(-b)}{\exp(\exp(-b))-1}=\exp(-b) \\
  b=-\log(\log(C)
  \end{aligned}
    \label{eq:weight_int_2}
\end{equation}
For the case of LVIS dataset that has 1,203 classes plus one for the background, we set the weights $W^T$ equal to $0.001$ and the bias equal to $-\log(\log(1204)\approx -2$. These values produce small initial gradients and they prevent gradient overflow.

\subsection{Temperature in Gumbel activation}
\begin{figure}[t]
    \centering
    \includegraphics[scale=0.29]{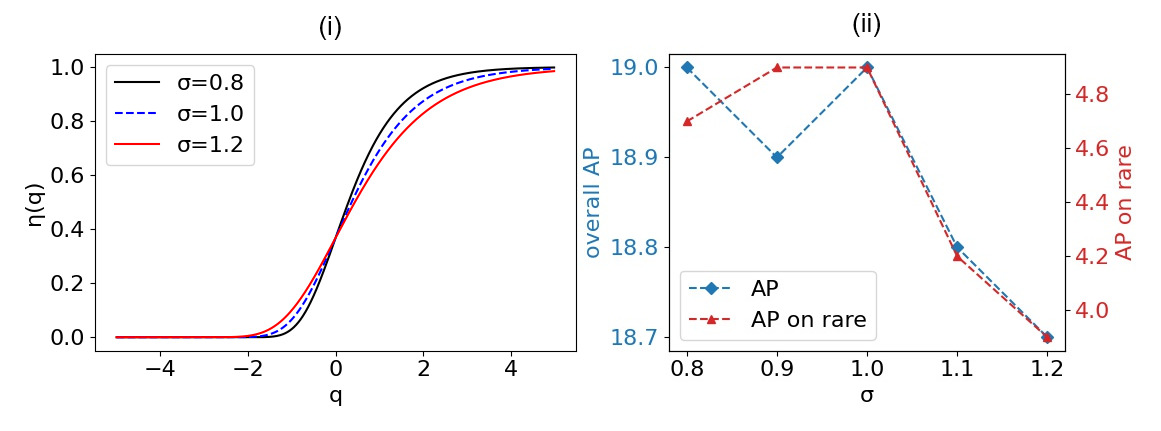}
    \caption{(i): Gumbel activation using different temperature $\sigma$. Selecting a larger $\sigma$ flattens the curve, while selecting a smaller $\sigma$ makes the curve steeper. (ii): Performance of MaskRCNN-R50 on LVISv1 using training schedule 1x and random sampler, for different choices of temperature $\sigma$. The best performance is observed for $\sigma=1.0$.}
    \label{fig:sigma_gumbel}
\end{figure}

We have also studied the choice of non Standard Gumbel activation, as shown in Figure \ref{fig:sigma_gumbel}.i, for different choices of temperature $\sigma$:
\begin{equation}
    \eta_{\gamma}(q_i;\sigma)=\exp(-\exp(-\frac{q_i}{\sigma}))
    \label{eq:sigm_gumbel}
\end{equation}
We observe that, choosing a larger temperature flattens Gumbel activation curve, while choosing a smaller temperature steepens the curve.
Gumbel activation has a double exponent as shown in Eq.~\ref{eq:sigm_gumbel}, which makes it difficult to select values of $\sigma$ due to arithmetic instability. In our case, we choose values $[0.8,0.9,1.0,1.1,1.2]$ and we observe that the best choice is  $\sigma=1$ as it has better overall $AP$ and $AP^r$ as shown in ~\ref{fig:sigma_gumbel}.ii.

\subsection{Gumbel activation and cut-off error}
Gumbel activation has a double exponent, as shown in Eq.~\ref{eq:sigm_gumbel}, this makes it numerically unstable for large inputs and hinders training. For this reason, we tested different ranges of values and decided to clip the input space to be within the range of $[-4,10]$. Using this range of values the cut-off error is \textit{e-5} and training commences without overflow errors. In the future, we will develop a solution that prevents numerical instability, so that we do not have to clip the input space.
\begin{figure}[t]
    \centering
    \includegraphics[scale=0.6]{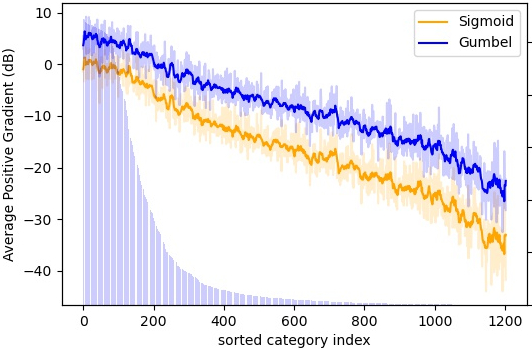}
    \caption{Average Positive Gradient $g$ per category, measured in decibel, (dB). Gumbel activation produces larger gradients for rare categories and facilitates rare category learning.}
    \label{fig:avg_pos_grad}
\end{figure}

\subsection{Average Positive Gradient}
We visualize the average positive gradient $g$, each category receives during training for 12 epochs using MaskRCNN. We use logarithmic scale to measure $g$ in $dB$ because the average gradient is small, especially for rare categories. As Figure \ref{fig:avg_pos_grad} indicates, using Gumbel activation, the positive gradient is on average $7 dB$ larger than the case of using Sigmoid, while for the case of rare categories, Gumbel produces gradients that are $10 dB$ larger.

In conclusion, the network learns better the rare categories by using Gumbel activation than by using Sigmoid activation, as the gradient is larger with Gumbel. This is also reflected in the formula of the positive Sigmoid gradient and the positive Gumbel gradient.
In detail, Sigmoid positive gradient is bounded to values $(-1,0)$, while Gumbel positive gradient is exponential and has values that reach $(-\infty , 0)$. This enables Gumbel activation to produce larger gradients than Sigmoid and it is useful for rare categories, where the gradient updates are scarce.

\subsection{Gumbel Optimised Loss}
Our $GOL$ method is based in DropLoss \cite{hsieh2021droploss}. It is described as follows:
\begin{equation}
    \mathcal{L}_{GOL} = -\sum_{j=1}^{C}w_j^{Drop}\log(\bar{p_j}),\;\; \bar{p_j}=\begin{cases}
         \eta_{\gamma}(q_i),\;\; if \;\;  y_j=1\\
         1-\eta_{\gamma}(q_i),\;\;  if\;\;  y_j=0
        \end{cases}
\end{equation}
\begin{equation}
    w_j^{Drop}=\begin{cases}
         1- T_{\lambda}(f_j)(1-y_j),\;\; if \;\;  E(r)=1\\
         w \sim \text{Ber}(\mu_{f_j}),\;\;  \;\;  otherwise
        \end{cases}
\end{equation}
\begin{equation}
    \mu_{f_j}=\begin{cases}
         (n_{rare}+n_{common})/n_{all},\;\; if \;\;  T_{\lambda}(f_j)=1\\
         n_{frequent}/n_{all},\;\;  \;\;  otherwise
        \end{cases}
\end{equation}
where $E(r)$ is a binary indicator function that outputs 1 if a region proposal $r$ is foreground, $T_{\lambda}(f_j)$ is a rare category indicator that outputs 1 if the frequency of category $j$ is lower than $\lambda$, $w \in \{0,1\}$ is a random variable drawn from Bernoulli distribution and $\mu_{f_j}$ is the shape parameter that is computed according to the foreground region proposals in the training batch.

\section{Long-tailed instance segmentation}
\label{sec:ltis_supp}
\subsection{Ablation Study}
\label{sec:ablation_studies}
In Table \ref{tab:detailed_ablation}, we conduct an ablation study of Gumbel activation, RFS \cite{gupta2019lvis}, EQL \cite{tan2020equalization}, DropLoss \cite{hsieh2021droploss}, Normalised Mask \cite{wang2021seesaw} and stricter Non Maximum Suppression (NMS) threshold. We denote the stricter NMS threshold and Normalised Mask enhancements as (Enh).

\begin{table}
    \centering
    \caption{Ablation study, using MaskRCNN, Resnet50 and training schedule 2x.}
    \begin{tabular}{ccccc|c|c|c|c|c}
         RFS&Gumbel&EQL&Enh&DropLoss&$AP$ &$AP^r$  &$AP^c$  &$AP^f$  &$AP^b$  \\
         \hline
         &&&&&18.7&1.1&16.2&29.2&19.5\\
          &\checkmark& & &&22.0&8.9&20.3&29.6&22.4\\
          &&\checkmark& &&21.6&3.8&21.7&29.2&22.5\\
          &\checkmark&\checkmark& &&23.9&11.4&23.4&29.9&24.2\\
         \checkmark&&&&&23.7&13.3&23.0&29.0&24.7\\
         \checkmark&\checkmark&&&&23.5&13.8&22.2&29.2&24.3\\
         \checkmark&&\checkmark &&&25.3&17.4&24.9&29.2&26.0\\
         \checkmark&\checkmark&\checkmark&&&26.1&18.4&25.9&29.8&26.8\\
         \hline
         \checkmark&\checkmark&\checkmark&\checkmark&&26.9&18.1&26.5&\textbf{31.3}&26.8\\
         \hline
         &\checkmark&&\checkmark&\checkmark&25.6&14.5&26.1&29.9&25.1\\
         \checkmark&\checkmark&&\checkmark&\checkmark&\textbf{27.7}&\textbf{21.4}&\textbf{27.7}&30.4&\textbf{27.5}\\

    \end{tabular}
    \label{tab:detailed_ablation}
\end{table}

As shown in Table \ref{tab:detailed_ablation} the best overall performance is achieved with Gumbel, RFS, Enh and DropLoss, we denote this pipeline as Gumbel Optimised Loss ($GOL$). The best performance on $AP^f$ is achieved using Gumbel, RFS, Enh and EQL, we denote this pipeline as $GOL$*. 

\subsubsection{Total Performance} Our $GOL$ method significantly boosts the vanilla MaskRCNN $AP$ by $9.0\%$, and it largely improves $AP^r$ by $20.3\%$, $AP^c$ by $11.5\%$, $AP^f$ by $1.2\%$ and $AP^b$ by $8.0\%$.

\begin{table}
    \centering
    \caption{MaskRCNN with Resnet50, schedule 1x, EQLv1 loss \cite{tan2020equalization}, DropLoss \cite{hsieh2021droploss}, ACSL \cite{wang2021adaptive} and Federated Loss \cite{zhou2021probablistic}. Gumbel activation boosts $AP$ of all models.}
    \begin{tabular}{c|c|c|c|c|c|c}
         Method&Activation&$AP$&$AP^r$&$AP^c$&$AP^f$&$AP^b$  \\
         \hline
         EQL$^\dagger$\cite{tan2021equalization}&Sigmoid& 18.6&2.1&17.4&27.2&19.3 \\
          EQL&Gumbel&\textbf{21.7}&\textbf{9.6}&\textbf{20.6}&\textbf{28.2}&\textbf{21.8}\\
         \hline
         DropLoss$^\dagger$\cite{hsieh2021droploss}&Sigmoid& 19.8&3.5&20.0&26.7&20.4 \\
         DropLoss&Gumbel&\textbf{22.0}&\textbf{10.0}&\textbf{22.1}&\textbf{27.1}&\textbf{21.9} \\
         \hline
         ACSL \cite{wang2021adaptive} &Sigmoid  &20.7  &9.6  &19.7 &26.6&\textbf{21.2}\\
         ACSL &Gumbel &\textbf{21.0} &\textbf{10.9}&\textbf{19.8}&\textbf{26.7}&21.1\\
         \hline
         Federated Loss \cite{zhou2021probablistic} &Sigmoid &17.6  &1.8 &14.9&27.5&18.2\\
         Federated Loss &Gumbel &\textbf{20.1}&\textbf{6.0}&\textbf{18.5}&\textbf{28.0}&\textbf{20.5}\\
    \end{tabular}
    \label{tab:detailed_gumbel_sota}
\end{table}

\begin{table}
    \centering
    \caption{Comparison of activations in various frameworks using 1x schedule.}
    \begin{tabular}{c|c|c|c|c|c|c}
         Method& Framework& $AP$ &$AP^r$  &$AP^c$  &$AP^f$  &$AP^b$  \\
         \hline
         Sigmoid& \multirow{ 3}{*}{MaskRCNN-ResNet50\cite{he2017mask}} &16.4 &0.8 &12.7 &27.3 &17.2 \\
         Softmax& &15.2 & 0.0& 10.6 &26.9&16.1\\
         Gumbel& & \textbf{19.0}&\textbf{4.9}&\textbf{16.8}&\textbf{27.6}&\textbf{19.1}\\
         \hline
         Sigmoid& \multirow{ 3}{*}{MaskRCNN-ResNet101} &17.8 &0.9 &14.5 &28.8 &18.8\\
         Softmax& &16.7 & 0.5& 12.5 &28.5&17.7\\
         Gumbel& & \textbf{20.6}&\textbf{6.4}&\textbf{18.5}&\textbf{29.2}&\textbf{21.0}\\
         \hline
         Sigmoid&\multirow{ 3}{*}{MaskRCNN-ResNeXt101} &19.6 &1.0&16.5 &31.2&20.7\\
         Softmax& &18.6 &0.6&14.5 &31.1&19.7\\
         Gumbel& & \textbf{22.6}&\textbf{5.9}&\textbf{21.3}&\textbf{31.4}&\textbf{22.8}\\
         \hline
         Sofmax&\multirow{ 2}{*}{Cascade MaskRCNN-Resnet101\cite{cai2019cascade}}  &18.8  &0.6 &15.7 &30.3 &21.3\\
         Gumbel& & \textbf{22.9} &\textbf{6.6}&\textbf{22.4}&\textbf{30.7}&\textbf{25.8}\\
         \hline
         Sofmax&\multirow{ 2}{*}{Hybrid Task Cascade-ResNet101\cite{chen2019hybrid}}  &19.1  &0.6  &15.8  &31.0 &21.1\\
         Gumbel& &\textbf{23.3}&\textbf{6.1} &\textbf{22.7}&\textbf{31.4}&\textbf{25.6}\\
    \end{tabular}
    \label{tab:detailed_larger_models}
\end{table}

\begin{table}
    \centering
    \caption{Comparative results on LVISv1 using MaskRCNN-FPN and schedule 2x.}
    \begin{tabular}{c|c|c|c|c|c|c|c}
         Method&Sampler&Backbone & $AP$&$AP^r$&$AP^c$&$AP^f$&$AP^b$ \\
         \hline
         Softmax&\multirow{6}{*}{random}&\multirow{6}{*}{MaskRCNN ResNet50}&18.7&1.1&16.2&29.2&19.5\\
         LOCE\cite{feng2021exploring}&&&23.8 &8.3 &23.7&30.7&24.0\\
         EQLv2\cite{tan2021equalization}&& &25.5&17.7&24.3&30.2&26.1\\
         Seesaw\cite{wang2021seesaw}&& &25.0 &16.1 &24.2&29.7&25.6\\
         Disalign\cite{zhang2021distribution}&&&24.2&13.2&23.8&29.3&24.7\\
         $GOL$ (ours)&&&25.6&14.5&26.1&29.9&25.1\\
         \hline
          LOCE[9]&MFS[9]&MaskRCNN ResNet50 &26.6 &18.5 &26.2&30.7&27.4\\
         \hline
         NorCal\cite{pan2021model}&\multirow{5}{*}{RFS}&\multirow{5}{*}{MaskRCNN ResNet50}&25.2&19.3&24.2&28.6&-\\
         EQLv2\cite{tan2021equalization}&&&25.8 &17.3 &25.4&30.0&26.2\\
         Seesaw\cite{wang2021seesaw}&&&26.4 &19.5&26.1&29.7&27.6\\
         $GOL$* (ours)& &&26.9&18.1&26.5&\textbf{31.3}&26.8\\
         $GOL$ (ours)&& &\textbf{27.7}&\textbf{21.4}&\textbf{27.7}&30.4&\textbf{27.5}\\
         \hline
         EQLv2\cite{tan2021equalization}&\multirow{3}{*}{random}&\multirow{3}{*}{MaskRCNN ResNet101}&27.2&20.6&25.9&31.4&27.9\\
         Seesaw\cite{wang2021seesaw}&&&27.1&18.7&26.3&31.7&27.4\\
         $GOL$ (ours)&&&27.0&16.1&27.4&31.2&26.8\\
         \hline
         LOCE[9]&MFS[9]&MaskRCNN ResNet101&28.0 &19.5 &27.8&32.0&29.0\\
         \hline
         NorCal\cite{pan2021model}&\multirow{4}{*}{RFS}&\multirow{4}{*}{MaskRCNN ResNet101}&27.3&20.8&26.5&31.0&28.1\\
         Seesaw\cite{wang2021seesaw}&&&28.1&20.0&28.0&31.8&28.9\\
         $GOL$*(ours)&&&28.0&19.3&27.5&\textbf{32.4}&28.3\\
         $GOL$(ours)&&&\textbf{29.0}&\textbf{22.8}&\textbf{29.0}&31.7&\textbf{29.2}\\
         
    \end{tabular}
    \label{tab:deep_models}
\end{table}

\subsection{Results on Larger Frameworks and SOTA Losses}
In Table \ref{tab:detailed_gumbel_sota}, we show detailed results when using Gumbel activation and SOTA long-tailed instance segmentation loss functions. In Table \ref{tab:detailed_larger_models}, we show detailed experimental results using Gumbel activation and common instance segmentation frameworks. In all cases, Gumbel activation improves the overall segmentation performance of models.

\subsection{Results on Larger Models}
We report the performance of our methods using larger models such as MaskRCNN with ResNet-101. As shown in Table \ref{tab:deep_models}, using MaskRCNN ResNet-50, $GOL$ significantly outperforms the best method, LOCE \cite{feng2021exploring} by $1.1\%$ on $AP$, by $2.9\%$ on $AP^r$ and by $1.5\%$ on $AP^c$, using smaller training budget and the same enhancements.

Using MaskRCNN ResNet-101, $GOL$ largely surpasses the best state-of-the-art Seesaw \cite{wang2021seesaw} by $0.9\%$ in overall $AP$, $2.8\%$ in $AP^r$, $1.0\%$ in $AP^c$ and $0.3\%$ in $AP^b$ using the same enhancements and RFS sampler. It also surpasses LOCE by $1.0\%$ in overall $AP$ using fewer training epochs.

Finally, our $GOL$* method has the best $AP^f$ in both MaskRCNN ResNet-50 and MaskRCNN ResNet-101 backbones, thus it is useful if $AP^f$ is most important.

\subsection{Object Distributions}
\begin{figure}[t]
    \centering
    \includegraphics[scale=0.45]{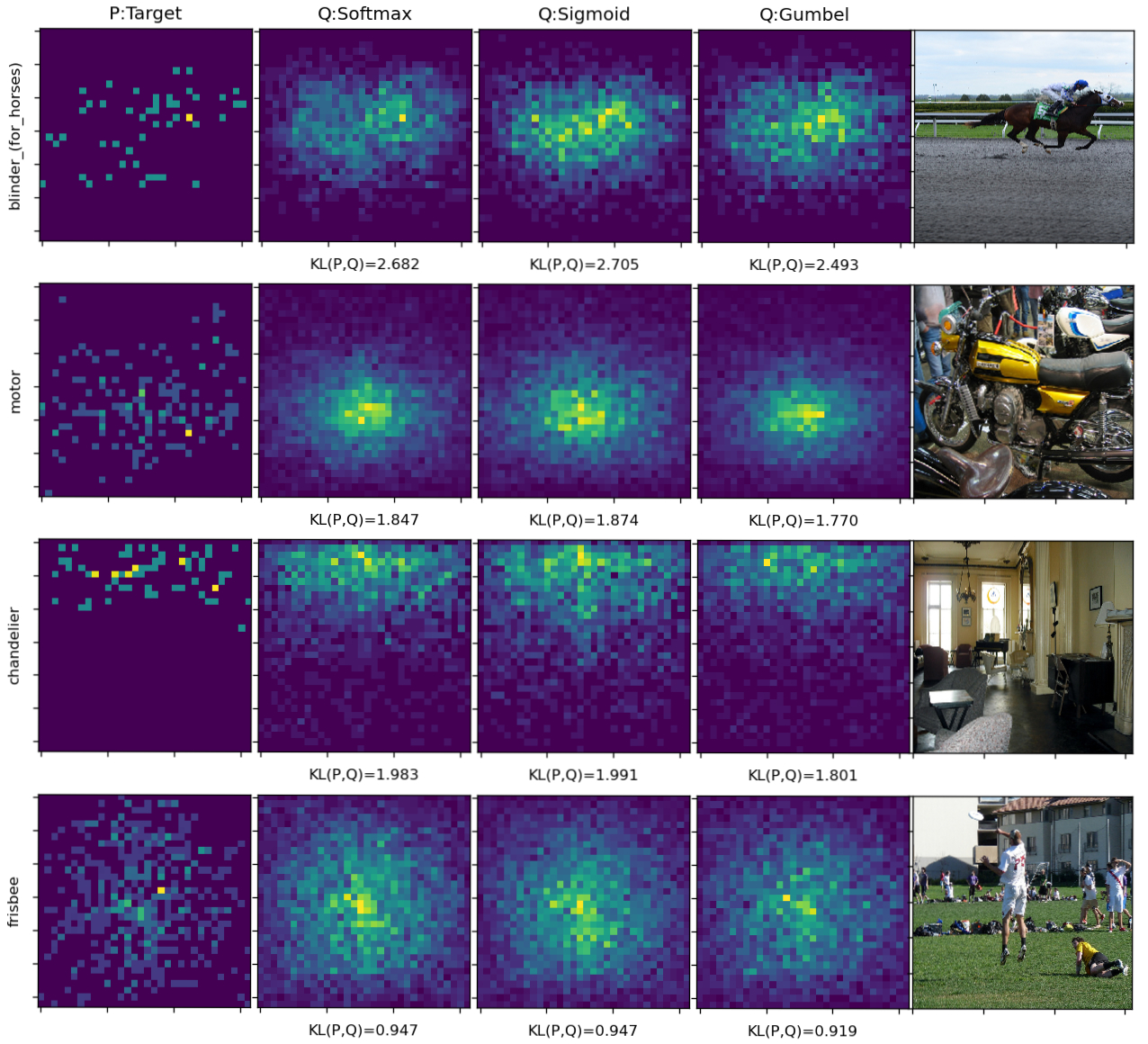}
    \caption{Comparison of four object distributions in LVIS validation set, using Softmax (second column), Sigmoid (third column) and Gumbel (fourth column). Gumbel predicts distributions that have smaller K-L divergence than Sigmoid or Softmax.}
    \label{fig:distance_4}
\end{figure}

We further show more examples of object distributions in LVIS v1 validation set. As shown in Figure \ref{fig:distance_4}, Gumbel activation produces object distributions that are closer to the target distribution as they have lower K-L divergence.

\begin{thebibliography}{10}
\providecommand{\url}[1]{\texttt{#1}}
\providecommand{\urlprefix}{URL }
\providecommand{\doi}[1]{https://doi.org/#1}

\bibitem{bridge2020introducing}
Bridge, J., Meng, Y., Zhao, Y., Du, Y., Zhao, M., Sun, R., Zheng, Y.:
  Introducing the gev activation function for highly unbalanced data to develop
  covid-19 diagnostic models. IEEE journal of Biomedical and Health Informatics
   \textbf{24}(10),  2776--2786 (2020)

\bibitem{cai2019cascade}
Cai, Z., Vasconcelos, N.: Cascade r-cnn: High quality object detection and
  instance segmentation. IEEE Transactions on Pattern Analysis and Machine
  Intelligence  (2019)

\bibitem{cao2019learning}
Cao, K., Wei, C., Gaidon, A., Arechiga, N., Ma, T.: Learning imbalanced
  datasets with label-distribution-aware margin loss. In: Advances in Neural
  Information Processing Systems (2019)

\bibitem{chawla2002smote}
Chawla, N.V., Bowyer, K.W., Hall, L.O., Kegelmeyer, W.P.: Smote: synthetic
  minority over-sampling technique. Journal of artificial intelligence research
   \textbf{16},  321--357 (2002)

\bibitem{chen2019hybrid}
Chen, K., Pang, J., Wang, J., Xiong, Y., Li, X., Sun, S., Feng, W., Liu, Z.,
  Shi, J., Ouyang, W., et~al.: Hybrid task cascade for instance segmentation.
  In: Proceedings of the IEEE/CVF Conference on Computer Vision and Pattern
  Recognition. pp. 4974--4983 (2019)

\bibitem{mmdetection}
Chen, K., Wang, J., Pang, J., Cao, Y., Xiong, Y., Li, X., Sun, S., Feng, W.,
  Liu, Z., Xu, J., Zhang, Z., Cheng, D., Zhu, C., Cheng, T., Zhao, Q., Li, B.,
  Lu, X., Zhu, R., Wu, Y., Dai, J., Wang, J., Shi, J., Ouyang, W., Loy, C.C.,
  Lin, D.: {MMDetection}: Open mmlab detection toolbox and benchmark. arXiv
  preprint arXiv:1906.07155  (2019)

\bibitem{cubuk2019autoaugment}
Cubuk, E.D., Zoph, B., Mane, D., Vasudevan, V., Le, Q.V.: Autoaugment: Learning
  augmentation strategies from data. In: Proceedings of the IEEE/CVF Conference
  on Computer Vision and Pattern Recognition. pp. 113--123 (2019)

\bibitem{cui2019class}
Cui, Y., Jia, M., Lin, T.Y., Song, Y., Belongie, S.: Class-balanced loss based
  on effective number of samples. In: Proceedings of the IEEE/CVF conference on
  computer vision and pattern recognition. pp. 9268--9277 (2019)

\bibitem{deng2009imagenet}
Deng, J., Dong, W., Socher, R., Li, L.J., Li, K., Fei-Fei, L.: Imagenet: A
  large-scale hierarchical image database. In: 2009 IEEE conference on computer
  vision and pattern recognition. pp. 248--255. Ieee (2009)

\bibitem{feng2021exploring}
Feng, C., Zhong, Y., Huang, W.: Exploring classification equilibrium in
  long-tailed object detection. In: Proceedings of the IEEE/CVF International
  Conference on Computer Vision. pp. 3417--3426 (2021)

\bibitem{gupta2019lvis}
Gupta, A., Dollar, P., Girshick, R.: Lvis: A dataset for large vocabulary
  instance segmentation. In: Proceedings of the IEEE/CVF Conference on Computer
  Vision and Pattern Recognition. pp. 5356--5364 (2019)

\bibitem{he2017mask}
He, K., Gkioxari, G., Doll{\'a}r, P., Girshick, R.: Mask r-cnn. In: Proceedings
  of the IEEE international conference on computer vision. pp. 2961--2969
  (2017)

\bibitem{he2016deep}
He, K., Zhang, X., Ren, S., Sun, J.: Deep residual learning for image
  recognition. In: Proceedings of the IEEE conference on computer vision and
  pattern recognition. pp. 770--778 (2016)

\bibitem{hsieh2021droploss}
Hsieh, T.I., Robb, E., Chen, H.T., Huang, J.B.: Droploss for long-tail instance
  segmentation. In: Proceedings of the AAAI conference on artificial
  intelligence. vol.~35, pp. 1549--1557 (2021)

\bibitem{hu2018squeeze}
Hu, J., Shen, L., Sun, G.: Squeeze-and-excitation networks. In: Proceedings of
  the IEEE conference on computer vision and pattern recognition. pp.
  7132--7141 (2018)

\bibitem{kang2019decoupling}
Kang, B., Xie, S., Rohrbach, M., Yan, Z., Gordo, A., Feng, J., Kalantidis, Y.:
  Decoupling representation and classifier for long-tailed recognition. In:
  Eighth International Conference on Learning Representations (ICLR) (2020)

\bibitem{khan2017cost}
Khan, S.H., Hayat, M., Bennamoun, M., Sohel, F.A., Togneri, R.: Cost-sensitive
  learning of deep feature representations from imbalanced data. IEEE
  transactions on neural networks and learning systems  \textbf{29}(8),
  3573--3587 (2017)

\bibitem{kim2020adjusting}
Kim, B., Kim, J.: Adjusting decision boundary for class imbalanced learning.
  IEEE Access  \textbf{8},  81674--81685 (2020)

\bibitem{kotz2000extreme}
Kotz, S., Nadarajah, S.: Extreme value distributions: theory and applications.
  World Scientific (2000)

\bibitem{krizhevsky2009learning}
Krizhevsky, A., Hinton, G., et~al.: Learning multiple layers of features from
  tiny images  (2009)

\bibitem{li2020overcoming}
Li, Y., Wang, T., Kang, B., Tang, S., Wang, C., Li, J., Feng, J.: Overcoming
  classifier imbalance for long-tail object detection with balanced group
  softmax. In: Proceedings of the IEEE/CVF conference on computer vision and
  pattern recognition. pp. 10991--11000 (2020)

\bibitem{lin2014microsoft}
Lin, T.Y., Maire, M., Belongie, S., Hays, J., Perona, P., Ramanan, D.,
  Doll{\'a}r, P., Zitnick, C.L.: Microsoft coco: Common objects in context. In:
  European conference on computer vision. pp. 740--755. Springer (2014)

\bibitem{liu2019large}
Liu, Z., Miao, Z., Zhan, X., Wang, J., Gong, B., Yu, S.X.: Large-scale
  long-tailed recognition in an open world. In: Proceedings of the IEEE/CVF
  Conference on Computer Vision and Pattern Recognition. pp. 2537--2546 (2019)

\bibitem{mahajan2018exploring}
Mahajan, D., Girshick, R., Ramanathan, V., He, K., Paluri, M., Li, Y.,
  Bharambe, A., Van Der~Maaten, L.: Exploring the limits of weakly supervised
  pretraining. In: Proceedings of the European conference on computer vision
  (ECCV). pp. 181--196 (2018)

\bibitem{menon2021longtail}
Menon, A.K., Jayasumana, S., Rawat, A.S., Jain, H., Veit, A., Kumar, S.:
  Long-tail learning via logit adjustment. In: International Conference on
  Learning Representations (2021),
  \url{https://openreview.net/forum?id=37nvvqkCo5}

\bibitem{mullapudi2021background}
Mullapudi, R.T., Poms, F., Mark, W.R., Ramanan, D., Fatahalian, K.: Background
  splitting: Finding rare classes in a sea of background. In: Proceedings of
  the IEEE/CVF Conference on Computer Vision and Pattern Recognition. pp.
  8043--8052 (2021)

\bibitem{oksuz2020imbalance}
Oksuz, K., Cam, B.C., Kalkan, S., Akbas, E.: Imbalance problems in object
  detection: A review. IEEE transactions on pattern analysis and machine
  intelligence  \textbf{43}(10),  3388--3415 (2020)

\bibitem{pan2021model}
Pan, T.Y., Zhang, C., Li, Y., Hu, H., Xuan, D., Changpinyo, S., Gong, B., Chao,
  W.L.: On model calibration for long-tailed object detection and instance
  segmentation. Advances in Neural Information Processing Systems  \textbf{34}
  (2021)

\bibitem{peng2020large}
Peng, J., Bu, X., Sun, M., Zhang, Z., Tan, T., Yan, J.: Large-scale object
  detection in the wild from imbalanced multi-labels. In: Proceedings of the
  IEEE/CVF Conference on Computer Vision and Pattern Recognition. pp.
  9709--9718 (2020)

\bibitem{Ren2020balms}
Ren, J., Yu, C., Sheng, S., Ma, X., Zhao, H., Yi, S., Li, H.: Balanced
  meta-softmax for long-tailed visual recognition. In: Proceedings of Neural
  Information Processing Systems(NeurIPS) (Dec 2020)

\bibitem{shen2016relay}
Shen, L., Lin, Z., Huang, Q.: Relay backpropagation for effective learning of
  deep convolutional neural networks. In: European conference on computer
  vision. pp. 467--482. Springer (2016)

\bibitem{tan2021equalization}
Tan, J., Lu, X., Zhang, G., Yin, C., Li, Q.: Equalization loss v2: A new
  gradient balance approach for long-tailed object detection. In: Proceedings
  of the IEEE/CVF Conference on Computer Vision and Pattern Recognition. pp.
  1685--1694 (2021)

\bibitem{tan2020equalization}
Tan, J., Wang, C., Li, B., Li, Q., Ouyang, W., Yin, C., Yan, J.: Equalization
  loss for long-tailed object recognition. In: Proceedings of the IEEE/CVF
  conference on computer vision and pattern recognition. pp. 11662--11671
  (2020)

\bibitem{tang2020long}
Tang, K., Huang, J., Zhang, H.: Long-tailed classification by keeping the good
  and removing the bad momentum causal effect. Advances in Neural Information
  Processing Systems  \textbf{33},  1513--1524 (2020)

\bibitem{wang2021seesaw}
Wang, J., Zhang, W., Zang, Y., Cao, Y., Pang, J., Gong, T., Chen, K., Liu, Z.,
  Loy, C.C., Lin, D.: Seesaw loss for long-tailed instance segmentation. In:
  Proceedings of the IEEE/CVF Conference on Computer Vision and Pattern
  Recognition. pp. 9695--9704 (2021)

\bibitem{wang2020devil}
Wang, T., Li, Y., Kang, B., Li, J., Liew, J., Tang, S., Hoi, S., Feng, J.: The
  devil is in classification: A simple framework for long-tail instance
  segmentation. In: European Conference on computer vision. pp. 728--744.
  Springer (2020)

\bibitem{wang2021adaptive}
Wang, T., Zhu, Y., Zhao, C., Zeng, W., Wang, J., Tang, M.: Adaptive class
  suppression loss for long-tail object detection. In: Proceedings of the
  IEEE/CVF Conference on Computer Vision and Pattern Recognition. pp.
  3103--3112 (2021)

\bibitem{wu2020forest}
Wu, J., Song, L., Wang, T., Zhang, Q., Yuan, J.: Forest r-cnn: Large-vocabulary
  long-tailed object detection and instance segmentation. In: Proceedings of
  the 28th ACM International Conference on Multimedia. pp. 1570--1578 (2020)

\bibitem{xie2017aggregated}
Xie, S., Girshick, R., Doll{\'a}r, P., Tu, Z., He, K.: Aggregated residual
  transformations for deep neural networks. In: Proceedings of the IEEE
  conference on computer vision and pattern recognition. pp. 1492--1500 (2017)

\bibitem{zhang2021distribution}
Zhang, S., Li, Z., Yan, S., He, X., Sun, J.: Distribution alignment: A unified
  framework for long-tail visual recognition. In: Proceedings of the IEEE/CVF
  Conference on Computer Vision and Pattern Recognition. pp. 2361--2370 (2021)

\bibitem{zhou2021probablistic}
Zhou, X., Koltun, V., Kr{\"a}henb{\"u}hl, P.: Probabilistic two-stage
  detection. In: arXiv preprint arXiv:2103.07461 (2021)

\bibitem{zou2018unsupervised}
Zou, Y., Yu, Z., Kumar, B., Wang, J.: Unsupervised domain adaptation for
  semantic segmentation via class-balanced self-training. In: Proceedings of
  the European conference on computer vision (ECCV). pp. 289--305 (2018)

\end{thebibliography}
\end{document}